\documentclass[11pt,a4paper]{article}
\usepackage[hyperref]{emnlp2020}
\usepackage{times}
\usepackage{latexsym}
\usepackage{booktabs}

\usepackage{microtype}

\aclfinalcopy % Uncomment this line for the final submission
\def\parcite#1{\citep{#1}} % (Smith, 2012)
\def\perscite#1{\citet{#1}} % Smith (2012)
\def\inparcite#1{\citealp{#1}} % should be Smith, 2012

\usepackage{amsmath}
\usepackage[capitalise]{cleveref}

\title{Announcing CzEng 2.0 Parallel Corpus with over 2 Gigawords}

\author{Tom Kocmi
        \qquad Martin Popel
        \qquad Ond{\v{r}}ej Bojar
        \\ \\
        Charles University, Faculty of Mathematics and Physics \\
        Institute of Formal and Applied Linguistics \\
        Malostransk{\'{e}} n{\'{a}}m{\v{e}}st{\'{\i}} 25, 118 00 Prague, Czech Republic \\
         {\tt \{kocmi,popel,bojar\}@ufal.mff.cuni.cz}}

\date{}

\begin{document}
\maketitle
\begin{abstract}
% \XXX{Nechame tohle poradi autoru nebo Kocmi, Bojar, Popel?}
%MP: Mně je to jedno. Někde se cení naopak poslední autor. Tak ať si vybere Ondra.

We present a new release of the Czech-English parallel corpus CzEng~2.0 consisting of over 2~billion words (2~``gigawords'') in each language.
The corpus contains document-level information and is filtered with several techniques to lower the amount of noise. In addition to the data in the previous version of CzEng, it contains new authentic and also high-quality synthetic parallel data. CzEng is freely available for research and educational purposes.

\end{abstract}

\section{Introduction}

This paper describes the new release of Czech-English parallel corpus CzEng~2.0. The version number is aligned with the year of the release, 2020.
CzEng~2.0 is the sixth release of the corpus and serves as a replacement for the previous version, CzEng~1.6 \parcite{czeng16}. There was also an intermediate release of CzEng~1.7 that filtered mostly noisy sentences out of CzEng~1.6. However, there was no accompanying publication.
In the newest release, we replicate some of the filterings of CzEng~1.7 with several additional.

The parallel corpus CzEng was successfully used in multiple NLP experiments, most notably in the WMT shared translation tasks since 2010, see \perscite{callisonburch-EtAl:2010:WMT} through \perscite{barrault2019wmt}. 

CzEng releases are freely available for research, and educational purposes and restricted versions of CzEng have been separately licensed for commercial use. 

When designing the current release, we aimed at the following goals:

\begin{itemize}
    \item providing document-level split,
    \item filtering noisy data,
    \item including new authentic data,
    \item generating high-quality synthetic data.
\end{itemize}

These goals are aligned with the latest development in Neural Machine Translation (NMT), where the quality and quantity of parallel data are one of the most critical parts for developing high-quality NMT systems. 

The corpus is available at \url{https://ufal.mff.cuni.cz/czeng/czeng20}.

This paper has the following structure: we describe sources of authentic data in \cref{sec:data_sources} and synthetic data as well as the process of their generation in \cref{sec:synthetic}. The filtering of parallel data is described in \cref{sec:filtering}. Information about the data format and IDs of sentences are in \cref{sec:data_format}. Lastly, we analyze the corpus, provide its statistics in \cref{sec:statistics} and conclude in \cref{sec:conclusion}.

\section{Data sources}
\label{sec:data_sources}

CzEng 2.0 contains restructured parallel data from the previous version of CzEng 1.6 \parcite{czeng16} and also new parallel data from various sources. In this section, we discuss each group separately.

Majority of authentic parallel sentences are from the preceding version of CzEng, which is segmented on a sentence-level with segments identification intact. The segments are usually short paragraphs or consecutive sentences from the original document. Each segment contains up to 15 sentences. Therefore, we could recreate the document-level information.

Additionally, we modified the distribution of data from CzEng 1.6, specifically the split between training, development and evaluation set. The development and evaluation sets have not been widely used, mainly as other official testsets are usually used to compare MT results such as WMT News testsets \parcite{barrault2019wmt}. Therefore, we have decided to merge the development set into the training part of the corpus and preserve only the evaluation set separated.

The second part of the parallel data comes from various new data sources. We have collected all Czech--English data from WMT 2020\footnote{\url{http://www.statmt.org/wmt20/translation-task.html}} and preprocessed them to follow the CzEng data format. New parallel data come from Europarl (v10), News commentary, Wikititles, Commoncrawl, Paracrawl\footnote{\url{https://www.paracrawl.eu/index.php}}, WikiMatrix \parcite{schwenk2019wikimatrix}, and Tilde MODEL Corpus (EESC, EMA, Rapid; \inparcite{rozis2017tilde}).

We analyzed new corpora manually and performed a pre-filtering on News commentary, Paracrawl, and Wikititles. CzEng 1.6 already contained News commentary. Therefore, to avoid duplicates, we have removed all sentences from the newer version of News commentary that is contained in CzEng 1.6. 
Paracrawl and Wikititles seemed highly noisy. Therefore, we removed all sentences where FastText \parcite{joulin2016fasttext} identified Czech or English with less than 50\% probability. This filtering is more strict than the one we applied to the completed corpus (see \cref{sec:filtering}).

Most of the new corpora are segmented on sentence-level except for Europarl and Rapid corpus, where we preserve the document-level segmentation for both of them. In the case of Europarl, we separated documents based on the speaker not based on whole sessions. This resulted in documents with an average length of 59 sentences, which is closer to original CzEng's document segments of length 15.

Furthermore, it is essential to mention that we have not added any new parallel data into the CzEng testset. Thus, its distribution of sentences no longer reflect the training data distribution. We have made this decision because the CzEng testset is not usually used for MT evaluation, and having larger training data is crucial for NMT.

\section{Synthetic backtranslated data}
\label{sec:synthetic}

We used the English-to-Czech and Czech-to-English models of \citet{popel2018WMT}  to translate monolingual English (``enmono'') and Czech (``csmono'') news crawl data provided by WMT\footnote{\url{http://data.statmt.org/news-crawl/}}
 and create thus synthetic parallel data.

All the source data is document level, and we kept the original document boundaries,  i.e. unlike in the authentic data from CzEng~1.6, there are documents longer than 15 sentences in the synthetic data.
However, the models used for backtranslation are sentence-level,  so the synthetic side of the data (Czech side in enmono, English side in csmono) may lack cross-sentence consistency.

The enmono data is a crawl from English news servers from 2016--2018,  resulting in 76M sentence pairs after filtering (see \cref{sec:filtering}).
The csmono data is a crawl from Czech news servers from 2013--2018,  resulting in 51M sentence pairs after filtering.

The models of \citet{popel2018WMT} were trained following the approach of \citet{popel-bojar:2018},  but with iterated backtranslation in a \emph{concat} regime,\footnote{
  The models were trained on CzEng~1.7 and WMT news crawl (English 2016--2017 and Czech 2007--2017),
   i.e. a subset of the sources listed in \cref{sec:data_sources}.
 } where the authentic and synthetic parallel data are simply concatenated (without shuffling), and last eight hourly checkpoints are averaged.
The models are transformer\_big trained in Tensor2Tensor \citep{vaswani2017attention}.
For decoding, we used beam size 4 and \texttt{alpha=1}. The only difference from the setup of \citet{popel2018WMT} is that we omitted the coreference preprocessing and regex post-processing.

\section{Filtering}
\label{sec:filtering}

Other authors and we noticed that CzEng~1.6 is noisy and needs further filtering \parcite{bojar-etal-2017-results-wmt17,popel2018WMT, bawden2019WMT}. There has been an effort to filter out noisy sentences from CzEng~1.6 released as CzEng~1.7, where 7\% training data have been removed. However, this effort has not been documented. Our filtering pipeline consists of recreating filtering for CzEng~1.7, followed by further filtering of all parallel sentences described in \cref{sec:data_sources}.

For CzEng~1.7, we apply document-level filtering, which makes the approach more conservative than sentence-level filtering. The first step is filtering corpus based on automatic language identification via Langid.py \parcite{lui-baldwin-2012-langid}. We drop all documents (segments of 15 sentences) where either the Czech or English side is recognized as a different language. Then we have removed documents where the Czech side did not contain any characters with Czech diacritics. This filtering should not remove many correct documents because in Czech, on average, almost every second word contains at least one accented character. Therefore the chance that a whole document would not contain any is minimal .\footnote{
This filter removes not only documents in other languages than Czech,  but also Czech documents are written without diacritics and various ``non-linguistic'' content, such as lists of football or stock-market results.
}

%udelal jsem si rychly vypocet nad NEWCORPORA casti czengu a vyslo mi ze 47.5% slov obsahuje diakritiku
Lastly, we have performed document-level deduplication removing identical documents.
These filtering techniques remove either all sentences in a given document or none. The filtering removed 4.1M sentence pairs from the training part of CzEng~1.6.
% \XXX{Tome, máš statistiky, kolik který filtr vyhodil vět?
% Tedy např. kolik vět/dokumentů po Langid.py bylo ještě vyhozeno kvůli absenci diakritiky?
% Osobně by mě to zajímalo, ale jestli to nemáš po ruce, tak to nehledej.}

The document-level filtering was followed by sentence-level filtering. It consists of removing extremely long sentences, removing sentence pairs based on automatic language identification, and dual conditional cross-entropy filtering \parcite{junczys-dowmunt-2018-dual}. We removed sentences based on quite conservative thresholds but provided computed scores in the final release for further filtering in tasks, where smaller, but cleaner data is needed.

First, we removed sentences longer than 200 (space-separated) words or 1600 characters. These are unnatural sentences, mostly containing lists of items or sentences that are incorrectly segmented.

Second, we used automatic language identification tool. In contrast to document-level filtering, we used FastText \parcite{joulin2016fasttext} because it has better accuracy on shorter texts. We computed language score separately for each language as follows:

\begin{equation}
\begin{split}
cs\_lang\_score = \frac{p(lang=Czech)}{p(lang=x)}\\
en\_lang\_score = \frac{p(lang=English)}{p(lang=x)}
\end{split}
\end{equation}

where $p$ are the probabilities assigned by FastText that $lang$ is a language of a given sentence and $x$ is the most probable language. In other words, it is a scaled probability that takes into account a situation when FastText is not sure about any of the languages and returns a similar probability for several languages -- a scenario typical for short sentences. Based on this score, we have removed sentence pairs with more than ten words in either language that also have $cs\_lang\_score$ or $en\_lang\_score$ lower than 0.5.

Third, we applied dual conditional cross-entropy filtering \citep{junczys-dowmunt-2018-dual}. It uses an NMT model to assign each sentence pair an adequacy score using conditional cross-entropy. The score is calculated as follows:

\begin{equation}
\begin{split}
crossent\_score=|H_A(y|x) - H_B(x|y)| \\
+\frac{1}{2}(H_A(y|x) + H_B(x|y)) \\
H_A = -log(P_A(en|cs)) \\
H_B = -log(P_B(cs|en)) \\
\end{split}
\end{equation}

where $H_A$ and $H_B$ are word-normalized conditional cross-entropies assigned by NMT models in one of translation directions.

The final score is negated and exponentiated,
 so that the values are between 0 (worst sentence pairs) and 1 (best):

\begin{equation}
\begin{split}
adq\_score = exp(-crossent\_score)
\end{split}
\end{equation}

We use models trained by \perscite{popel2018WMT} to compute cross-entropies.\footnote{
While \citet{junczys-dowmunt-2018-dual} trained the scoring models on ``small subsamples of clean data'',  we used the best models available to us,  i.e. models trained on all the data (authentic and synthetic). This means filtering the data by using scoring models trained on the same data.
We checked manually that most of the sentence pairs with low $adq\_score$ are noisy and should be filtered.
However, we noticed that the sentence pairs with the highest $adq\_score$ were often long sentences duplicated many times in the training data (note that we performed document deduplication, but not sentence deduplication).
Thus taking e.g. top 5\% of the data will not give optimal results.
For the future works, we suggest to do sentence-level deduplication before training the scoring models, but just document (or paragraph) deduplication for the final filtering.
}
These models won WMT 2018 MT in both directions for Czech--English language pair \parcite{bojar-EtAl:2018:WMT1} and should be good at scoring sentences. Based on our manual examination, we have removed all sentence pairs that obtained $adq\_score$ less than 0.02.

Our filtering steps removed only a small part of the corpus that is the noisiest because we expect researchers to apply further filtering of their own.
We have added our scores into the final corpus, so it is easy to select a smaller and cleaner corpus based on the scores.
The IDs also contain the source name, so it is possible to filter out the noisiest (or most out-of-domain for a given purpose) sources, e.g. Subtitles, Paracrawl and WikiMatrix.

\section{Corpus Data Format}
\label{sec:data_format}

CzEng is shuffled on a document level, and empty lines separate individual documents.

\def\FN#1{\textsf{#1}}
The final corpus contains four files: \FN{train}, \FN{test}, \FN{csmono} and \FN{enmono}. \FN{train} contains all the authentic parallel training data.
\FN{test} is a filtered version of the `evaluation set' from CzEng~1.6 with authentic parallel data.
The \FN{csmono} and \FN{enmono} files are the synthetic parallel data (cf. \cref{sec:synthetic}).

Each file contains six tab-separated columns: unique ID, adq\_score, cs\_lang\_score, en\_lang\_score, Czech sentence, English sentence.
All three scores are within 0 and 1, and higher values mean better scores (cleaner sentence pairs).
For the synthetic data, none of the three scores can be reliably computed, so all the three scores are set to 1.

Each sentence pair was assigned a unique ID containing the data source, document ID, file ID and sentence pair ID. We use the ID system from CzEng~1.6 and extend it to all new data.
For example, \texttt{paracrawl-b16598886-f0-s1} specifies that a given sentence pair comes from the Paracrawl corpus, it is from a document with ID=b16598886-f0, and it is the first sentence in the document.

\section{Corpus Analysis}
\label{sec:statistics}

The final number of sentences, number of Czech and English words in our corpus is showed in \cref{tab:final_sizes}.
In contrast to the previous version of CzEng~1.6, the new release contains 9M new sentence pairs, but we removed 10M noisy sentences.
Therefore, the authentic part of the corpora has roughly the same amount of parallel sentences as in Czeng~1.6.
Interestingly, it has slightly more words, which is mainly due to newly added corpora that have, on average, more words per sentence.

The synthetic part of the corpus contains 51M parallel sentences generated from Czech monolingual data and 76M parallel sentences from English monolingual data.

In total, the corpus contains 2.6 Czech gigawords and 3.0 English gigawords.

\begin{table}[t]\centering\small
\begin{tabular}{@{}l@{~~}rrr@{}}\toprule
Description           & Sent. pairs & CS words & EN words \\\midrule
CzEng 1.6             & 62 M & 611 M & 689 M \\ %62493539 & 611094657 & 688534368 \\
CzEng 1.7             & 57 M & 546 M & 622 M \\ %57383670 546244473 621915229
New corpora           & 9 M & 162 M & 183 M \\\midrule
Auth. filtered part   & 61 M & 617 M & 702 M\\
Synth. from Czech     & 51 M & 700 M & 833 M\\
Synth. from English   & 76 M & 1296 M & 1474 M\\
Test set              & 0.5 M & 4 M & 5 M\\\midrule
Final CzEng 2.0       & 188 M & 2618 M & 3013 M\\ % 188301322 2617896752 3013422886
\bottomrule
\end{tabular}
\caption{Statistics of number of sentences, Czech words and English words (space separated). Top part of the table presents previous versions of CzEng and sizes of newly added corpora before filtering. Middle part represents sizes of individual filtered CzEng 2.0 parts. Last row is a total size of CzEng 2.0 altogether.}
\label{tab:final_sizes}
\end{table}
%   92748 commoncrawl
%  5677929 fiction
%  1291384 eesc
%   480464 ema
%  6163127 eu
%   643448 europarl
%   859874 medical
%   29528 navajo
%   241357 news
%   68132 newscommentary
%  5052622 paracrawl
%   481573 paraweb
%   302014 pdfs
%   438942 rapiddoc
%  1430955 techdoc
%      501 tweets
%  1123891 wikimatrix
%   193973 wikititles
% 36408183 subtitles*

\subsection{Machine Translation Experiment}

In order to test the primary goal of the trainset, improving the quality of machine translation, we train baseline with various sizes of the CzEng corpus to measure the performance.
We use the Tensor2Tensor framework \parcite{tensor2tensor} and the architecture Transformer-big as described by \perscite{vaswani2017attention}.
Each model is trained for 1M training steps with a batch size of 4500 subwords on two GPUs. We use Adafactor as the optimizer and inverse square-root learning rate with 16k warm-up steps.
The vocabulary is identical for all models and has a size of 32k subwords.
We use checkpoint averaging over the last four checkpoints distanced by 25000 steps.
During the inference, we use and beam size of 8 and alpha 0.8. We should mention that these systems have lower quality than \perscite{popel2018WMT}.

The final results are measured on the English-to-Czech concatenated test set from years 2012--2019 \parcite{barrault2019wmt} with case sensitive SacreBLEU \parcite{post2018sacrebleu}.\footnote{SaceBLEU signature: \texttt{BLEU+case.mixed+numrefs.1 +smooth.exp+tok.13a+version.1.4.6}} 
We computed performance separately for testset sentences originated in Czech (in total 6854 sentences), and sentences originated in English (in total 10936 sentences). These sentences are selected by SacreBLEU option \texttt{--origlang}. The results are presented in \cref{tab:bleu_scores}. In our analysis, we focus on translation ``orig-EN''. This it is a more realistic setting as human translators also translated these sentences from English into (translationese) Czech.

Unfiltered CzEng 2.0 has 0.3 BLEU worse performance compared to filtered version. This confirms that our filtering helped and we removed mostly noisy sentences. 
Another interesting observation is that the performance stays almost the same when reducing corpus size based on $adq\_score$ score shows, except for the situation when we keep only 16.1M training sentences.

Additionally, whenever we mix the ``enmono'' synthetic data into the training corpus, we get additional improvements of 1.6 BLEU. The highest performance is obtained by training on all of the parallel data leading to improvements of 1.7 BLEU.

On the other hand, adding ``csmono'' sentences lowered performance when evaluated on ``orig-EN'' but improved performance on ``orig-CS''. We think that this could be because the model could learn to generate more natural-looking sentences than translationese Czech that is tested in ``orig-EN''. However, this should be investigated more in-depth in future works.

% \XXX{asi odstranit sloupecek BEAM1 a DEV, byl kvuli overeni}
% \begin{table*}[t]
% \centering
% \small
% \begin{tabular}{lrrrrrr}
% Trainset           & Sent. pairs & BEAM 8 & BEAM 1 & DEV & orig CS & orig EN \\
% \hline
% Unfiltered CzEng2.0    &  69.0 M & 23.9 & 23.6 & 26.4 & 26.3 & 26.4\\
% CzEng2.0               &  60.9 M & 24.2 & 24.0 & 26.5 & 26.6 & 26.8\\
% Filter 0.1             &  50.5 M & 24.2 & 24.0 & 26.5 & 26.5 & 26.9\\
% Filter 0.25            &  34.5 M & 24.2 & 23.9 & ?    & 26.1 & 26.7\\
% Filter 0.5             &  16.1 M & 22.3 & 22.0 & 24.2 & 24.1 & 24.6\\
% CE20 + csmono          & 111.6 M & 25.8 & 25.6 & 25.1 & 29.6 & 26.2\\
% CE20 + enmono          & 137.2 M & 25.8 & 25.7 & 27.6 & 28.2 & 28.0\\
% CE20 + csmono + enmono & 187.8 M & 26.1 & 26.0 & 27.1 & 29.3 & 28.0\\
% \end{tabular}
% \caption{Evaluated experiment for model translating from English into Czech. The rows ``Filter'' define new threshold for $adq\_score$ score based on which we reduce the size of training corpus.}
% \label{tab:bleu_scores}
% \end{table*}

\begin{table}[t]\centering\small
\begin{tabular}{@{}l@{~}r@{~~}r@{~~}r@{}}\toprule
                                       &             & \multicolumn{2}{c}{BLEU}\\\cmidrule{3-4}
Train set                              & Sent. pairs & orig CS & orig EN \\\midrule
Unfiltered CzEng2.0 \FN{train}         &  69.0 M & 26.8 & 27.1\\
CzEng2.0 \FN{train}                    &  60.9 M & 27.2 & 27.4\\
Filter 0.1 \FN{train}                  &  50.5 M & 26.9 & 27.4 \\
Filter 0.25 \FN{train}                 &  34.5 M & 26.7 & 27.2 \\
Filter 0.5 \FN{train}                  &  16.1 M & 24.3 & 25.0\\
\FN{train} + \FN{csmono}               & 111.6 M & 30.1 & 26.4 \\
\FN{train} + \FN{enmono}               & 137.2 M & 28.4 & 28.0 \\
\FN{train} + \FN{csmono} + \FN{enmono} & 187.8 M & 29.7 & 28.1\\\bottomrule
\end{tabular}
\caption{BLEU evaluation of our English-to-Czech experiments.
The rows ``Filter'' define an alternative threshold for $adq\_score$, based on which we reduce the size of training corpus.}
\label{tab:bleu_scores}
\end{table}

\section{Conclusion}
\label{sec:conclusion}

We introduced a new release of the Czech--English parallel corpus CzEng, version 2.0. We hope that the new release will follow the success and popularity of the previous versions.
CzEng 2.0 is enlarged, contains new authentic and high-quality synthetic parallel data. We removed the noisiest parts and included filtering scores for further cleaning. We especially highlight the document level segmentation, which we believe is necessary for further development of machine translation.
We noticed that the corpus contains many nearly identical sentences, so for future work, we plan experiments with filtering these near duplicates.
% could be used to improve performance. On the other hand, this could be an advantage of the corpora.

% \section{Unused Notes}

% normalizace podle delky nam kazila distribuci viz excel

% neni tam moc near duplicates soudime podle ze :
% Tak to je zajímavé: v CzEngu máme 674080 duplicitních vět delších než 20 slov,
% ale pouze 636 z nich má právě jeden možný překlad do druhého jazyka
% (a ten překlad není překladem žádné jiné věty).
% Přičemž jako duplicitní beru druhé až další opakování téže věty.

\bibliography{emnlp2020}

\begin{thebibliography}{16}
\expandafter\ifx\csname natexlab\endcsname\relax\def\natexlab#1{#1}\fi

\bibitem[{Barrault et~al.(2019)Barrault, Bojar, Costa-jussa, Federmann, Fishel,
  Graham, Haddow, Huck, Koehn, Malmasi, Monz, Muller, Pal, Post, and
  Zampieri}]{barrault2019wmt}
Loac Barrault, Ondrej Bojar, Marta~R. Costa-jussa, Christian Federmann, Mark
  Fishel, Yvette Graham, Barry Haddow, Matthias Huck, Philipp Koehn, Shervin
  Malmasi, Christof Monz, Mathias Muller, Santanu Pal, Matt Post, and Marcos
  Zampieri. 2019.
\newblock \href {http://www.aclweb.org/anthology/W19-5301} {Findings of the
  2019 conference on machine translation (wmt19)}.
\newblock In \emph{Proceedings of the Fourth Conference on Machine Translation
  (Volume 2: Shared Task Papers, Day 1)}, pages 1--61, Florence, Italy.
  Association for Computational Linguistics.

\bibitem[{Bawden et~al.(2019)Bawden, Bogoychev, Germann, Grundkiewicz, Kirefu,
  Miceli~Barone, and Birch}]{bawden2019WMT}
Rachel Bawden, Nikolay Bogoychev, Ulrich Germann, Roman Grundkiewicz, Faheem
  Kirefu, Antonio~Valerio Miceli~Barone, and Alexandra Birch. 2019.
\newblock \href {http://www.aclweb.org/anthology/W19-5304} {{The University of
  Edinburgh's Submissions to the WMT19 News Translation Task}}.
\newblock In \emph{Proceedings of the Fourth Conference on Machine Translation
  (Volume 2: Shared Task Papers, Day 1)}, pages 103--115, Florence, Italy.
  Association for Computational Linguistics.

\bibitem[{Bojar et~al.(2016)Bojar, Du{\v{s}}ek, Kocmi, Libovick{\'{y}},
  Nov{\'{a}}k, Popel, Sudarikov, and Vari{\v{s}}}]{czeng16}
Ond{\v{r}}ej Bojar, Ond{\v{r}}ej Du{\v{s}}ek, Tom Kocmi, Jind{\v{r}}ich
  Libovick{\'{y}}, Michal Nov{\'{a}}k, Martin Popel, Roman Sudarikov, and
  Du{\v{s}}an Vari{\v{s}}. 2016.
\newblock {CzEng 1.6: Enlarged Czech-English Parallel Corpus with Processing
  Tools Dockered}.
\newblock In \emph{{Text, Speech, and Dialogue: 19th International Conference,
  {TSD} 2016}}, number 9924 in Lecture Notes in Computer Science, pages
  231--238, Cham / Heidelberg / New York / Dordrecht / London. Masaryk
  University, Springer International Publishing.

\bibitem[{Bojar et~al.(2018)Bojar, Federmann, Fishel, Graham, Haddow, Huck,
  Koehn, and Monz}]{bojar-EtAl:2018:WMT1}
Ondrej Bojar, Christian Federmann, Mark Fishel, Yvette Graham, Barry Haddow,
  Matthias Huck, Philipp Koehn, and Christof Monz. 2018.
\newblock \href {http://www.aclweb.org/anthology/W18-6401} {{Findings of the
  2018 Conference on Machine Translation (WMT18)}}.
\newblock In \emph{Proceedings of the Third Conference on Machine Translation,
  Volume 2: Shared Task Papers}, pages 272--307, Belgium, Brussels. Association
  for Computational Linguistics.

\bibitem[{Bojar et~al.(2017)Bojar, Helcl, Kocmi, Libovick{\'y}, and
  Musil}]{bojar-etal-2017-results-wmt17}
Ond{\v{r}}ej Bojar, Jind{\v{r}}ich Helcl, Tom Kocmi, Jind{\v{r}}ich
  Libovick{\'y}, and Tom{\'a}{\v{s}} Musil. 2017.
\newblock \href {https://doi.org/10.18653/v1/W17-4757} {Results of the {WMT}17
  neural {MT} training task}.
\newblock In \emph{Proceedings of the Second Conference on Machine
  Translation}, pages 525--533, Copenhagen, Denmark. Association for
  Computational Linguistics.

\bibitem[{Callison-Burch et~al.(2010)Callison-Burch, Koehn, Monz, Peterson,
  Przybocki, and Zaidan}]{callisonburch-EtAl:2010:WMT}
Chris Callison-Burch, Philipp Koehn, Christof Monz, Kay Peterson, Mark
  Przybocki, and Omar Zaidan. 2010.
\newblock \href {http://www.aclweb.org/anthology/W10-1703} {{Findings of the
  2010 Joint Workshop on Statistical Machine Translation and Metrics for
  Machine Translation}}.
\newblock In \emph{Proceedings of the Joint Fifth Workshop on Statistical
  Machine Translation and MetricsMATR}, pages 17--53, Uppsala, Sweden.
  Association for Computational Linguistics.
\newblock Revised August 2010.

\bibitem[{Joulin et~al.(2016)Joulin, Grave, Bojanowski, Douze, J{\'e}gou, and
  Mikolov}]{joulin2016fasttext}
Armand Joulin, Edouard Grave, Piotr Bojanowski, Matthijs Douze, H{\'e}rve
  J{\'e}gou, and Tomas Mikolov. 2016.
\newblock Fasttext.zip: Compressing text classification models.
\newblock \emph{arXiv preprint arXiv:1612.03651}.

\bibitem[{Junczys-Dowmunt(2018)}]{junczys-dowmunt-2018-dual}
Marcin Junczys-Dowmunt. 2018.
\newblock \href {https://doi.org/10.18653/v1/W18-6478} {Dual conditional
  cross-entropy filtering of noisy parallel corpora}.
\newblock In \emph{Proceedings of the Third Conference on Machine Translation:
  Shared Task Papers}, pages 888--895, Belgium, Brussels. Association for
  Computational Linguistics.

\bibitem[{Lui and Baldwin(2012)}]{lui-baldwin-2012-langid}
Marco Lui and Timothy Baldwin. 2012.
\newblock \href {https://www.aclweb.org/anthology/P12-3005} {langid.py: An
  off-the-shelf language identification tool}.
\newblock In \emph{Proceedings of the {ACL} 2012 System Demonstrations}, pages
  25--30, Jeju Island, Korea. Association for Computational Linguistics.

\bibitem[{Popel(2018)}]{popel2018WMT}
Martin Popel. 2018.
\newblock \href {http://www.aclweb.org/anthology/W18-6424} {{CUNI Transformer
  Neural MT System for WMT18}}.
\newblock In \emph{Proceedings of the Third Conference on Machine Translation,
  Volume 2: Shared Task Papers}, pages 486--491, Belgium, Brussels. Association
  for Computational Linguistics.

\bibitem[{Popel and Bojar(2018)}]{popel-bojar:2018}
Martin Popel and Ond{\v{r}}ej Bojar. 2018.
\newblock \href {https://doi.org/10.2478/pralin-2018-0002} {{Training Tips for
  the Transformer Model}}.
\newblock \emph{The Prague Bulletin of Mathematical Linguistics}, 110:43--70.

\bibitem[{Post(2018)}]{post2018sacrebleu}
Matt Post. 2018.
\newblock \href {https://www.aclweb.org/anthology/W18-6319} {{A Call for
  Clarity in Reporting BLEU Scores}}.
\newblock In \emph{Proceedings of the Third Conference on Machine Translation:
  Research Papers}, pages 186--191, Belgium, Brussels. Association for
  Computational Linguistics.

\bibitem[{Rozis and Skadin{\v{s}}(2017)}]{rozis2017tilde}
Roberts Rozis and Raivis Skadin{\v{s}}. 2017.
\newblock Tilde model-multilingual open data for eu languages.
\newblock In \emph{Proceedings of the 21st Nordic Conference on Computational
  Linguistics, NoDaLiDa, 22-24 May 2017, Gothenburg, Sweden}, 131, pages
  263--265. Link{\"o}ping University Electronic Press.

\bibitem[{Schwenk et~al.(2019)Schwenk, Chaudhary, Sun, Gong, and
  Guzm{\'a}n}]{schwenk2019wikimatrix}
Holger Schwenk, Vishrav Chaudhary, Shuo Sun, Hongyu Gong, and Francisco
  Guzm{\'a}n. 2019.
\newblock Wikimatrix: Mining 135m parallel sentences in 1620 language pairs
  from wikipedia.
\newblock \emph{arXiv preprint arXiv:1907.05791}.

\bibitem[{Vaswani et~al.(2018)Vaswani, Bengio, Brevdo, Chollet, Gomez, Gouws,
  Jones, Kaiser, Kalchbrenner, Parmar, Sepassi, Shazeer, and
  Uszkoreit}]{tensor2tensor}
Ashish Vaswani, Samy Bengio, Eugene Brevdo, Francois Chollet, Aidan Gomez,
  Stephan Gouws, Llion Jones, Lukasz Kaiser, Nal Kalchbrenner, Niki Parmar,
  Ryan Sepassi, Noam Shazeer, and Jakob Uszkoreit. 2018.
\newblock \href {http://www.aclweb.org/anthology/W18-1819} {{Tensor2Tensor for
  Neural Machine Translation}}.
\newblock In \emph{Proceedings of the 13th Conference of the Association for
  Machine Translation in the Americas (Volume 1: Research Papers)}, pages
  193--199, Boston, MA. Association for Machine Translation in the Americas.

\bibitem[{Vaswani et~al.(2017)Vaswani, Shazeer, Parmar, Uszkoreit, Jones,
  Gomez, Kaiser, and Polosukhin}]{vaswani2017attention}
Ashish Vaswani, Noam Shazeer, Niki Parmar, Jakob Uszkoreit, Llion Jones,
  Aidan~N Gomez, {\L}ukasz Kaiser, and Illia Polosukhin. 2017.
\newblock \href
  {http://papers.nips.cc/paper/7181-attention-is-all-you-need.pdf} {{Attention
  is All you Need}}.
\newblock In I.~Guyon, U.~V. Luxburg, S.~Bengio, H.~Wallach, R.~Fergus,
  S.~Vishwanathan, and R.~Garnett, editors, \emph{Advances in Neural
  Information Processing Systems 30}, pages 6000--6010. Curran Associates, Inc.

\end{thebibliography}
\bibliographystyle{acl_natbib}

\end{document}